\documentclass[10pt,twocolumn,letterpaper]{article}

\usepackage{iccv}              
\usepackage[accsupp]{axessibility}
\usepackage{url}
\usepackage{booktabs}
\usepackage{algorithm}
\usepackage{algorithmic}
\usepackage{bbding}
\usepackage{multirow}
\usepackage{amsmath}
\usepackage{amssymb}
\usepackage{colortbl}
\usepackage{amsfonts} 
\usepackage{mathtools}
\usepackage{amsthm}
\usepackage{array}
\usepackage{xcolor}
\definecolor{darkgreen}{rgb}{0.0, 0.5, 0.0}
\usepackage{subcaption}
\usepackage{enumitem}
\usepackage{arydshln}
\usepackage{microtype}
\usepackage{epsfig}
\usepackage{bm}
\usepackage{bbding}
\usepackage{pifont}

\usepackage{dsfont}
\usepackage{color}
\definecolor{iccvblue}{rgb}{0.21,0.49,0.74}
\usepackage[pagebackref,breaklinks,colorlinks,allcolors=iccvblue]{hyperref}

\def\paperID{6523} 
\def\confName{ICCV}
\def\confYear{2025}

\title{SAM2Long: Enhancing SAM 2 for Long Video Segmentation \\ with a Training-Free Memory Tree}
\author{
    \textbf{Shuangrui Ding}$^{1}$ ~
    \textbf{Rui Qian}$^{1}$ ~
    \textbf{Xiaoyi Dong}$^{1,2}$\thanks{Corresponding Author.} ~
    \textbf{Pan Zhang}$^2$ ~ \\
    \textbf{Yuhang Zang}$^2$ ~
    \textbf{Yuhang Cao}$^2$ ~
    \textbf{Yuwei Guo}$^{1}$
    \textbf{Dahua Lin}$^{1,2,3}$ ~
    \textbf{Jiaqi Wang}$^{2}$\footnotemark[1]\\
    $^1$\ The Chinese University of Hong Kong 
    $^2$\ Shanghai AI Laboratory 
    $^3$\ CPII under InnoHK\\
    {\tt\small https://mark12ding.github.io/project/SAM2Long/}
}

\begin{document}
\maketitle

\begin{abstract}
The Segment Anything Model 2 (SAM~2) has emerged as a powerful foundation model for object segmentation in both images and videos.
The crucial design of SAM~2 for video segmentation is its memory module, which prompts object-aware memories from previous frames for current frame prediction.
However, its greedy-selection memory design suffers from the ``error accumulation" problem, where an errored or missed mask will cascade and influence the segmentation of the subsequent frames, which limits the performance of SAM~2 toward complex long-term videos.
To this end, we introduce SAM2Long, an improved \textbf{training-free} video object segmentation strategy, which considers the segmentation uncertainty within each frame and chooses the video-level optimal results from multiple segmentation pathways in a constrained tree search manner. 
In practice, we maintain a fixed number of segmentation pathways throughout the video.
For each frame, multiple masks are proposed based on the existing pathways, creating various candidate branches. We then select the same fixed number of branches with higher cumulative scores as the new pathways for the next frame. After processing the final frame, the pathway with the highest cumulative score is chosen as the final segmentation result.
Benefiting from its heuristic search design, SAM2Long is robust toward occlusions and object reappearances, and can effectively segment and track objects for complex long-term videos. 
Without further training, SAM2Long significantly and consistently outperforms SAM~2 on nine VOS benchmarks and three VOT benchmarks. Notably, SAM2Long achieves an average improvement of \textbf{3.7} points across all 12 direct comparisons, with gains of up to \textbf{5.3} points in $\mathcal{J} \& \mathcal{F}$ on long-term video object segmentation benchmarks such as SA-V and LVOS. The code is released at \url{https://github.com/Mark12Ding/SAM2Long}. 
\end{abstract}


\section{Introduction}
The Segment Anything Model 2 (SAM~2) has gained significant attention as a unified foundational model for promptable object segmentation in both images and videos. Notably, SAM~2~\citep{ravi2024sam} has achieved state-of-the-art performance across various video object segmentation tasks, significantly surpassing previous methods. Building upon the original SAM~\citep{kirillov2023segment}, SAM~2 incorporates a memory module that enables it to generate masklet predictions using stored memory contexts from previously observed frames. This module allows SAM~2 to seamlessly extend SAM into the video domain, processing video frames sequentially, attending to the prior memories of the target object, and maintaining object coherence over time.

\begin{figure*}[t]
    \centering
    \includegraphics[width=0.99\linewidth]{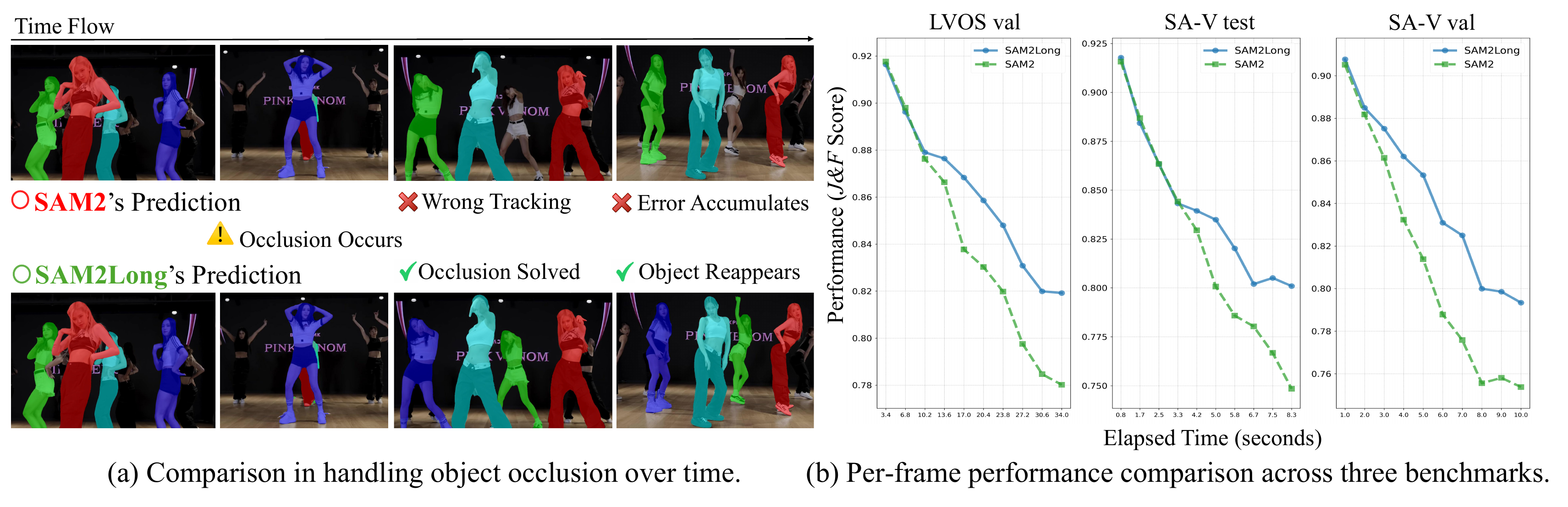}
    \caption{Comparison of occlusion handling and long-term capability between SAM~2 and SAM2Long. (a) When an occlusion occurs, SAM~2 may lose track or follow the wrong object, leading to accumulated errors. In contrast, SAM2Long utilizes memory tree search to recover when the object reappears. (b) Per-frame $\mathcal{J}\&\mathcal{F}$ scores of the predicted masks with different timestamp are plotted on the LVOS and SA-V datasets. SAM2Long demonstrates greater resilience to long video compared to SAM~2.}
    \label{fig:teaser}
    \vspace{-5pt}
\end{figure*}

While SAM~2 demonstrates strong performance in video segmentation and downstream tasks~\cite{guo2025keyframe}, its greedy segmentation strategy struggles to handle complex video scenarios with frequent occlusions and object reappearance. 
In detail, SAM~2 confidently and accurately segments frames when clear visual cues are present. However, in scenarios with occlusions or reappearing objects, it can produce mask proposals that are highly variable and uncertain.
Regardless of the frame's complexity, a uniform greedy selection strategy is applied to both scenarios: the mask with the highest predicted IoU is selected. 
Such greedy choice works well for the easy cases but raises the error potential for the challenging frames. Once an incorrect mask is selected into memory, it is uncorrectable and will mislead the segmentation of the subsequent frames. 
Figure~\ref{fig:teaser} illustrates the problem of ``error accumulation'' both qualitatively and quantitatively. The performance of SAM~2 gradually declines as the propagation progresses into later temporal segments, underscoring its limitations in keeping accurate tracking over time.

To this end, we redesign the memory module of SAM~2 to enhance its long-term capability and robustness against occlusions and error propagation. 
Our approach is motivated by the observation that the SAM~2 mask decoder generates multiple diverse masks, accompanied by predicted IoU scores and an occlusion score when handling challenging and ambiguous cases.
However, SAM~2 only selects a single mask as memory, sometimes disregarding the correct one. 
To address this limitation, we propose integrating SAM 2 with multiple memory pathways, inspired by prior Multiple Hypothesis Tracking (MHT) techniques~\citep{cox1996efficient, kim2015multiple, reid1979algorithm}. These techniques, widely adopted in the tracking community, enable the storage of various masks as memory at each time step and delay the association decision to obtain video-level optimal result.
In particular, we maintain a fixed number of memory pathways over time to explore multiple segmentation hypotheses with efficiently managed computational resources. At each time step, based on a set of memory pathways, each with its own memory bank and cumulative score (accumulated logarithm of the predicted IoU scores across the pathway), we produce multiple candidate branches for the current frame. 
Then, among all the branches, we only keep the same number of branches with higher cumulative scores and prune other branches, thereby constraining the tree's growth. After processing the final frame, the pathway with the highest cumulative score is selected as the final segmentation result. Moreover, to prevent premature convergence on incorrect predictions, we select pathways with distinct predicted masks when their occlusion scores indicate uncertainty. In this way, we maintain diversity in the tree branches in the challenging and distraction scenario. This tree-like memory structure augments SAM~2's ability to effectively overcome error accumulation. 

Within each pathway, we construct an object-aware memory bank that selectively includes frames with confidently detected objects and high-quality segmentation masks, based on the predicted occlusion scores and IoU scores. Instead of simply storing the nearest frames as SAM~2 does, we filter out frames where the object may be occluded or poorly segmented. This ensures that the memory bank provides effective object cues for the current frame's segmentation. Additionally, we modulate the memory attention calculation by weighting memory entries according to their occlusion scores, emphasizing more reliable entries during cross-attention. These strategies help SAM~2 focus on reliable object clues and improve segmentation accuracy with negligible computational overhead. As evidenced in Figure~\ref{fig:teaser}(a), our approach successfully resolves occlusions and re-tracks the recurring dancers, where SAM~2 fails. 

Our improvement is completely free of additional training and does not introduce any external parameters, but simply unleashes the potential of SAM~2 itself.
We provide a comprehensive evaluation that SAM2Long consistently outperforms SAM~2 across nine VOS benchmarks and three VOT benchmarks, particularly excelling in long-term and occlusion-heavy scenarios. Moreover, datasets with longer video durations generally lead to greater performance gains, which aligns with the motivation of our design. For instance, on the challenging SA-V test set, SAM2Long-L improves the $\mathcal{J}\&\mathcal{F}$ score by 5.3 points, and SAM2Long-S shows an impressive 4.7-point gain over the same size SAM~2 model on SA-V val set. Similar trends are observed on the LVOS validation set, where SAM2Long-S surpasses SAM 2-S by 3.5 points. These consistent improvements across different model sizes, including both SAM 2 and the more recent SAM 2.1 model weights, clearly demonstrate the generalization ability of our proposed method.
Furthermore, as shown in Figures~\ref{fig:teaser}(b), the per-frame performance gap between SAM2Long and SAM 2 increases over time. This highlights SAM2Long’s strength in long-term tracking scenarios.
With these results, we believe SAM2Long sets a new standard for video object segmentation based on SAM~2 in complex, real-world applications.

\section{Related work}
\subsection{Video Object Segmentation} 
Perceiving the environment in terms of objects is a fundamental cognitive ability of humans. In computer vision, Video Object Segmentation (VOS) tasks aim to replicate this capability by requiring models to segment and track specified objects within video sequences. 
A substantial amount of research has been conducted on video object segmentation in recent decades~\citep{fan2019shifting, oh2019video, hu2018motion, oh2018fast,perazzi2017learning,wang2019fast, hu2018videomatch, li2018video, bao2018cnn, zhang2019fast, li2020fast, johnander2019generative,zhang2023joint, ventura2019rvos, li2022recurrent,wu2023scalable,wang2023look, qian2023semantics, qian2024rethinking, ding2022motion, ding2023betrayed}.
There are two main protocols for evaluating VOS models~\citep{pont20172017,perazzi2016benchmark}: semi-supervised and unsupervised video object segmentation. In semi-supervised VOS, the first-frame mask of the objects of interest is provided, and the model tracks these objects in subsequent frames. In unsupervised VOS, the model directly segments the most salient objects from the background without any reference. It is important to note that these protocols are defined in the inference phase, and VOS methods can leverage ground truth annotations during the training stage. 
In this paper, we explore SAM~2~\citep{ravi2024sam}, for its application in semi-supervised VOS. 
We enhance the memory design of SAM~2, significantly improving mask propagation performance without any additional training.

\subsection{Memory-Based VOS}
Video object segmentation remains an unsolved challenge due to the inherent complexity of video scenes. Objects in videos can undergo deformation~\citep{tokmakov2023breaking}, exhibit dynamic motion~\citep{brox2010object}, reappear over long durations~\citep{hong2024lvos, hong2023lvos}, and experience occlusion~\citep{MOSE}, among other challenges.
To address the above challenges, adopting a memory architecture to store the object information from past frames is indispensable for accurately tracking objects in video~\citep{bhat2020learning, caelles2017one, maninis2018video, robinson2020learning, voigtlaender2017online, chen2018blazingly, hu2018videomatch, voigtlaender2019feelvos, yang2018efficient, yang2020collaborative, yang2021collaborative}. 
Recent approaches have introduced efficient memory reading mechanisms, utilizing either pixel-level attention~\citep{cheng2023tracking, zhou2024rmem, duke2021sstvos, liang2020video, oh2018fast, seong2020kernelized, cheng2022xmem, xie2021efficient, yang2022decoupling, yang2021associating} or object-level attention~\citep{athar2023tarvis, athar2022hodor, cheng2024putting}. 
Among these, Multiple Hypothesis Tracking (MHT)~\citep{cox1996efficient, kim2015multiple, reid1979algorithm} stands out as a robust paradigm that maintains tree search to construct and manage memory.
A prominent example is XMem~\citep{cheng2022xmem}, which leverages a hierarchical memory structure for pixel-level memory reading combined. Building on XMem's framework, Cutie~\citep{cheng2024putting} further improves segmentation accuracy by processing pixel features at the object level to better handle complex scenarios. 
The latest SAM~2~\citep{ravi2024sam} incorporates a simple memory module on top of the image-based SAM~\citep{kirillov2023segment}, enabling it to function for VOS tasks. 
However, SAM~2 struggles with challenging cases involving long-term reappearing objects and confusingly similar objects. 
Motivated by MHT, we redesign SAM~2's memory to maintain multiple potential correct masks, making the model more object-aware and robust.



\section{Method}
\subsection{Preliminary on SAM 2}
SAM~2~\citep{ravi2024sam} begins with an image encoder that encodes each input frame into embeddings. In contrast to SAM, where frame embeddings are fed directly into the mask decoder, SAM~2 incorporates a memory module that conditions the current frame’s features on both previous and prompted frames. Specifically, for semi-supervised video object segmentation tasks, SAM~2 maintains a memory bank at each time step $t \geq 1$:
$$
\mathcal{M}_t = \left\{ \mathbf{M}_{\tau} \in \mathbb{R}^{K \times C} \right\}_{\tau \in \mathcal{I}},
$$
where \( K \) is the number of memory tokens per frame, \( C \) is the channel dimension, and \( \mathcal{I} \) is the set of frame indices included in the memory. In SAM~2, memory set $\mathcal{I}$ stores up to \( N \) of the most recent frames, along with the initial mask, using a First-In-First-Out (FIFO) queue mechanism.

Each memory entry consists of two components: (1) the spatial embedding fused with the predicted mask (generated by the memory encoder), and (2) the object-level pointer (generated by the mask decoder). After cross-attending to the memory, the current frame's features integrate both fine-grained correspondences and object-level semantic information.\footnote{In practice, SAM~2 stores more object pointers than spatial embeddings, as pointers are lighter. We assume equal numbers of both components solely for illustrative purposes, without altering the actual implementation.}

The mask decoder, which is lightweight and retains the efficiency of SAM, then generates three predicted masks for the current frame. Each mask is accompanied by a predicted Intersection over Union (IoU) score $\text{IoU}_t \geq 0$ and an output mask token. 
Additionally, the mask decoder predicts a single occlusion score $o_t$ for the frame, where $o_t > 0$ indicates object presence, $o_t < 0$ indicates absence, and the absolute value $|o_t|$ depicts the model's confidence. The mask with the highest predicted IoU score is selected as the final prediction, and its corresponding output token is transformed into the object pointer for use as the memory.

\subsection{Constrained Tree Memory with Uncertainty Handling}

\begin{figure*}
    \centering    
    \includegraphics[width=\linewidth]{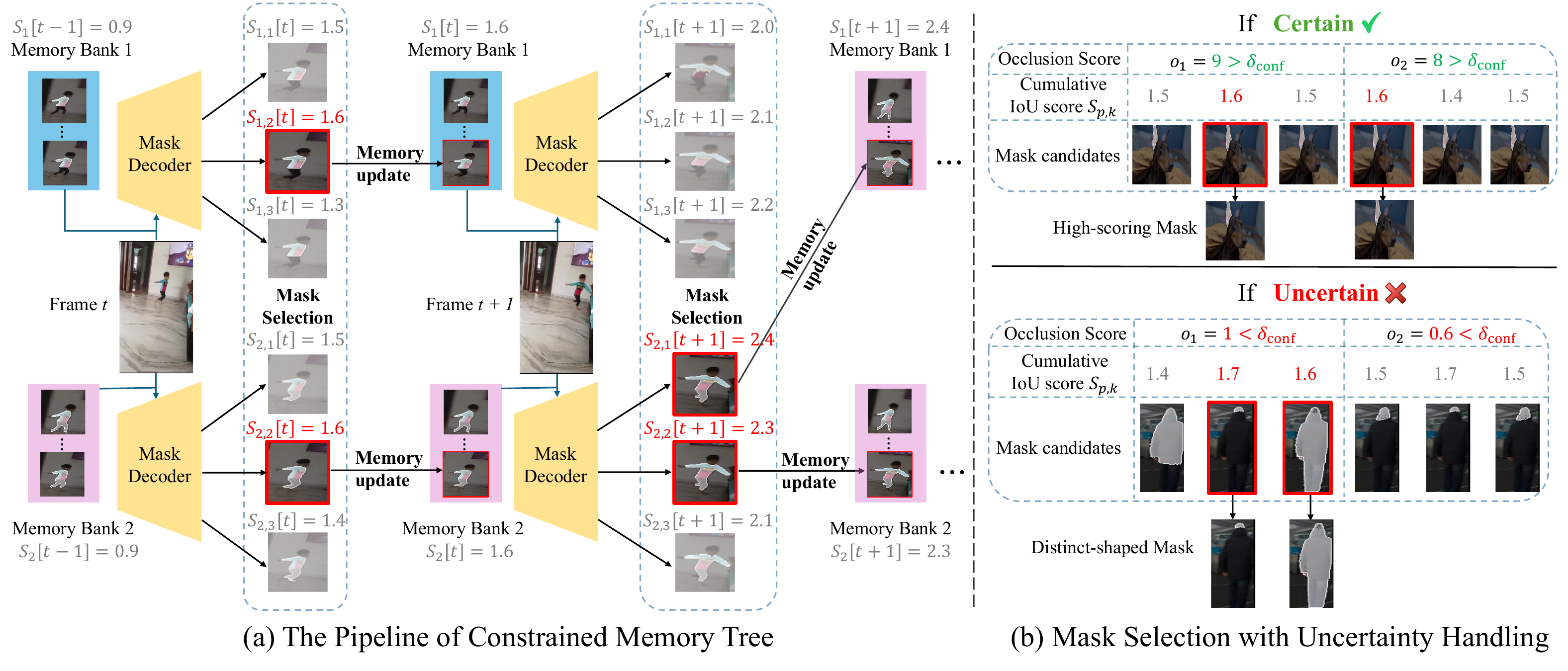}
    \caption{(a) The pipeline of constrained memory tree: At each time step $t$, we maintain multiple memory pathways, each containing a memory bank and a cumulative score \( S_p[t] \). The input frame is processed through the mask decoder conditioned on the memory bank, generating three mask candidates for each pathway. The candidates with the highest updated cumulative scores \( S_{p,k}[t] \) are carried forward to the next time step. (b) Mask selection with uncertainty handling: When the maximum absolute occlusion score exceeds the threshold $\delta_{\text{conf}}$ ({\textcolor{darkgreen}{Certain}}), the high-scoring mask is selected. Otherwise ({\textcolor{red}{Uncertain}}), distinct mask candidates are picked to avoid incorrect convergence.}
    \label{fig:pipeline}
    \vspace{-5pt}
\end{figure*}

To enhance SAM~2's robustness towards long-term and ambiguous cases, we propose a constrained tree memory structure that enables the model to explore various object states over time with minimal computational overhead. 
We show the high-level pipeline in Figure~\ref{fig:pipeline}. Note that we store memory separately for each target. Each target maintains its own individual tree memory bank. This tree-based approach maintains multiple plausible pathways and mitigates the effects of erroneous predictions. 

Specifically, at each time step \( t \), we maintain a set of \( P \) memory pathways, each with a memory bank \( \mathcal{M}_t^{p} \) and a cumulative score \( S_p[t] \), representing a possible segmentation hypothesis up to frame \( t \). Conditioned on the memory bank of each pathway \( p \), the SAM~2 decoder head generates three mask candidates along with their predicted IoU scores, denoted as \( \text{IoU}_t^{p,1} \), \( \text{IoU}_t^{p,2} \), and \( \text{IoU}_t^{p,3} \). 
This process expands the tree by branching each existing pathway into three new candidates. 
As a result, there are a total of \( 3P \) possible pathways at each time step.
We then calculate the cumulative scores for each possible pathway by adding the logarithm of its IoU score to the pathway's previous score:
\[
S_{p,k}[t] = S_p[t-1] + \log(\text{IoU}_t^{p,k} + \epsilon), \quad \text{for } k = 1, 2, 3,
\]
where \( \epsilon \) is a small constant to prevent the logarithm of zero. 

However, continuously tripling the pathways would lead to unacceptable computational and memory costs. 
Therefore, to manage computational complexity and memory usage, we implement a pruning strategy that selects the top \( P \) pathways with the highest cumulative scores to carry forward to the next time step. 
This selection not only retains the most promising segmentation hypotheses but also constrains the tree-based memory, ensuring computational efficiency.
Finally, we output the segmentation pathway with the highest cumulative score as the ultimate result.

Compared to SAM~2, our approach introduces additional computation mainly by increasing the number of passes through the mask decoder and memory module. Notably, these components are lightweight relative to the image encoder. For instance, the image encoder of SAM 2-Large consists of $212$M parameters while the total parameter of SAM 2-Large is $224$M. Since we process the image encoder only once just as SAM~2 does, the introduction of a memory tree adds negligible computational cost while significantly enhancing SAM~2's robustness against error-prone cases.

\begin{table*}[]
\centering
\small
\begin{tabular}{p{2.3cm} p{1.4cm} p{0.5cm} p{0.6cm} | p{1.4cm} p{0.5cm} p{0.6cm} | p{1.4cm} p{0.5cm} p{0.5cm}}
\toprule
\multirow{2}{*}{\textbf{Method}} & \multicolumn{3}{c}{\textbf{SA-V val}} & \multicolumn{3}{c}{\textbf{SA-V test}} & \multicolumn{3}{c}{\textbf{LVOS v2 val}}   \\
 & $\mathcal{J}\&\mathcal{F}$ & $\mathcal{J}$ & $\mathcal{F}$ & $\mathcal{J}\&\mathcal{F}$ & $\mathcal{J}$ & $\mathcal{F}$ & $\mathcal{J}\&\mathcal{F}$ & $\mathcal{J}$ & $\mathcal{F}$  \\
\hline
SAM2-T$^\dagger$ & 73.5 & 70.1 & 76.9 & 74.6 &  71.1& 78.0 & 77.8 & 74.5& 81.2\\
SAM2Long-T & 77.0 (\textcolor{red}{3.5$\uparrow$})& 73.2& 80.7 &78.7 (\textcolor{red}{4.1$\uparrow$}) & 74.6 & 82.7 & 81.4 (\textcolor{red}{3.6$\uparrow$}) & 77.7 & 85.0 \\ 
SAM2.1-T$^\dagger$ & 75.1 & 71.6 & 78.6 & 76.3 & 72.7 & 79.8 & 81.6 & 77.9 & 85.2 \\
SAM2.1Long-T & 78.9 (\textcolor{red}{3.8$\uparrow$})& 75.2 & 82.7 &79.0 (\textcolor{red}{2.7$\uparrow$}) & 75.2 & 82.9 & 82.4 (\textcolor{red}{0.8$\uparrow$}) & 78.8 & 85.9  \\ \hdashline
SAM2-S$^\dagger$ & 73.0  & 69.7 & 76.3 & 74.6 & 71.0 & 78.1 & 79.7 & 76.2 & 83.3\\
SAM2Long-S & 77.7 (\textcolor{red}{4.7$\uparrow$})& 73.9& 81.5 &78.1 (\textcolor{red}{3.5$\uparrow$})& 74.1& 82.0 & 83.2 (\textcolor{red}{3.5$\uparrow$}) & 79.5 & 86.8 \\ 
SAM2.1-S$^\dagger$ & 76.9 & 73.5 & 80.3 & 76.9 & 73.3 & 80.5& 82.1 & 78.6 & 85.6\\
SAM2.1Long-S & 79.6 (\textcolor{red}{2.7$\uparrow$}) & 75.9 & 83.3  & 80.4 (\textcolor{red}{3.5$\uparrow$})& 76.6& 84.1 & 84.3 (\textcolor{red}{2.2$\uparrow$}) & 80.7 & 88.0  \\ \hdashline
SAM2-B+$^\dagger$ & 75.4 & 71.9&  78.8 & 74.6 & 71.2 & 78.1 & 80.2 & 76.8 & 83.6\\
SAM2Long-B+ & 78.4 (\textcolor{red}{3.0$\uparrow$})& 74.7& 82.1 &78.5 (\textcolor{red}{3.9$\uparrow$}) & 74.7& 82.2 & 82.3 (\textcolor{red}{2.1$\uparrow$}) & 78.8 & 85.9  \\ 
SAM2.1-B+$^\dagger$ & 78.0 & 74.6 & 81.5 & 77.7 & 74.2 & 81.2 & 83.1 & 79.6 & 86.5 \\ 
SAM2.1Long-B+ & 80.5 (\textcolor{red}{2.5$\uparrow$})& 76.8 & 84.2 &80.8 (\textcolor{red}{3.1$\uparrow$}) & 77.1& 84.5 & 85.2 (\textcolor{red}{2.1$\uparrow$}) & 81.5 & 88.9\\ \hdashline
SAM2-L$^\dagger$ & 76.3& 73.0& 79.5 & 75.5 & 72.2 & 78.9 &83.0 &79.6& 86.4 \\
SAM2Long-L & 80.8 (\textcolor{red}{4.5$\uparrow$})& 77.1 & 84.5 & 80.8 (\textcolor{red}{5.3$\uparrow$}) & 76.8& 84.7 & 85.4 (\textcolor{red}{2.4$\uparrow$}) & 81.8 & 88.7  \\ 
SAM2.1-L$^\dagger$ & 78.6 & 75.1 & 82.0 & 79.6 & 76.1 & 83.2 & 84.0 & 80.7 &87.4 \\
SAM2.1Long-L & 81.1 (\textcolor{red}{2.5$\uparrow$})& 77.5& 84.7 & 81.2 (\textcolor{red}{1.6$\uparrow$}) & 77.6 & 84.9 & 85.3 (\textcolor{red}{1.3$\uparrow$}) &81.9 &88.8 \\ 
\bottomrule
\end{tabular}
\caption{Performance comparison on SA-V~\citep{ravi2024sam} and LVOS v2~\citep{hong2024lvos} datasets between SAM~2 and SAM2Long across all model sizes. $\dagger$ We report the re-produced performance of SAM~2(2.1) using its open-source code and checkpoint.}
\label{tab:main-results}
\end{table*}

\noindent\textbf{Uncertainty Handling.} Unfortunately, there are times when all pathways are uncertain. To prevent the model from improperly converging on incorrect predictions, we implement a strategy to maintain diversity among the pathways by deliberately selecting distinct masks.
That is, if the maximum absolute occlusion score across all pathways at time $t$, $\max(\{|o_t^p|\}_{p=1}^P)$, is less than a predefined uncertainty threshold $\delta_{\text{conf}}$, we enforce the model to select mask candidates with unique IoU values. This design is inspired by the observation that, within the same frame, different IoU scores typically correspond to distinct masks, which we verify in Table~\ref{tab:rounding_places}. In practice, we round each IoU score \( \text{IoU}_t^{p,k} \) to two decimal places and only select those hypotheses with distinct rounded values.

Overall, the integration of constrained tree memory with uncertainty handling offers a balanced strategy that leverages multiple segmentation hypotheses to enhance robustness toward the long-term complex video and achieve more accurate and reliable segmentation performance by effectively mitigating error accumulation.

\subsection{Object-aware Memory Bank Construction}
In each memory pathway, we devise object-aware memory selection to retrieve frames with discriminative objects. Meanwhile, we modulate the memory attention calculation to further strengthen the model's focus on the target objects. 

\vspace{2pt}
\noindent\textbf{Memory Frame Selection.}
To construct a memory bank that provides effective object cues, we selectively choose frames from previous time steps based on the predicted object presence and segmentation quality. Starting from the frame immediately before the current frame \( t \), we iterate backward through the prior frames \( i = \{t - 1, t - 2, \dots, 1\} \) in sequence. For each frame \( i \), we retrieve its predicted occlusion score \( o_i \) and IoU score \( \text{IoU}_i \) as reference. We include frame \( i \) in the memory bank if it satisfies the following criteria: 
\vspace{-0.2em}
\[\text{IoU}_i > \delta_{\text{IoU}} \quad \text{and} \quad o_i > 0,
\]
\vspace{-0.2em}
where \( \delta_{\text{IoU}} \) is a predefined IoU threshold. This ensures that only frames with confidently detected objects and reasonable segmentation masks contribute to the memory. We continue this process until we have selected up to \( N \) frames. In contrast to SAM~2, which directly picks the nearest \( N \) frames as the memory entries, this selection process effectively filters out frames where the object may be occluded, absent, or poorly segmented, thereby providing more robust object cues for the segmentation of the current frame.

\vspace{2pt}
\noindent\textbf{Memory Attention Modulation.}
To further emphasize more reliable memory entries during the cross-attention computation, we utilize the associated occlusion score $o_t$ to modulate the contribution of each memory entry. Assuming the memory set consists of $N$ frames plus the initial frame, totaling $N+1$ masks,
we define a set of standard weights \( \mathcal{W}^{\text{std}} \) that are linearly spaced between a lower bound \( w_{\text{low}} \) and an upper bound \( w_{\text{high}} \):
\[
\mathcal{W}^{\text{std}} = \left\{ w_{\text{low}} + \frac{i - 1}{N} (w_{\text{high}} - w_{\text{low}}) \right\}_{i=1}^{N+1}.
\]
Next, we sort the occlusion scores in ascending order to obtain sorted indices \( \mathcal{I'} = \{ I_i \}_{i=1}^{N+1} \) such that:
\[
o_{I_1} \leq o_{I_2} \leq \dots \leq o_{I_{N+1}}.
\]
We then assign the standard weights to the memory entries based on these sorted indices:
\[
w_{I_i} = \mathcal{W}^{\text{std}}_i, \quad \text{for } i = 1, 2, \dots, N+1.
\]
This assignment ensures that memory entries with higher occlusion scores, which indicate object presence with higher confidence, receive higher weights.
Then, we linearly scale the original keys \( \mathbf{M}_\tau \) with their corresponding weights:
$$
\widetilde{\mathbf{M}}_{\tau} = w_{\tau} \cdot \mathbf{M}_\tau, \quad \text{for } \tau \in \mathcal{I}.
$$
Finally, the modulated memory keys \( \widetilde{\mathcal{M}}_t  = \{ \widetilde{\mathbf{M}}_{\tau} \}_{\tau \in \mathcal{I}} \) are used in the memory module's cross-attention mechanism to update the current frame's features.
By using the available occlusion scores as indicators, we effectively emphasize memory entries with more reliable object cues.

\section{Experiments}
\subsection{Experiments Setup}
\noindent\textbf{Datasets.}
To evaluate our method, we select nine standard VOS benchmarks: SA-V val and test~\cite{ravi2024sam}, LVOS v1~\cite{hong2023lvos}, LVOS v2~\cite{hong2024lvos}, MOSE~\cite{MOSE}, VOST~\cite{tokmakov2023breaking}, PUMaVOS~\cite{bekuzarov2023xmem++}, DAVIS~\cite{pont20172017}, and YTVOS~\cite{xu2018youtube}. We report the following metrics: $\mathcal{J}$ (region similarity), $\mathcal{F}$ (contour accuracy), and the combined $\mathcal{J} \& \mathcal{F}$. All evaluations are conducted in a semi-supervised setting, where the first-frame mask is provided. Details on the benchmarks are provided in the Appendix.

\noindent\textbf{Evaluation.} 
SAM2Long is modified from the official codebase of SAM2 and we report the re-produced performance of SAM2 with the same official code and checkpoint. We noticed a slight difference between the re-produced results and the performance reported in the SAM2 paper, but for a strictly fair comparison, we tend to report both the SAM2 and SAM2Long results under the same hardware/software environment and settings. 
We set the uncertainty threshold to $\delta_{\text{conf}} = 2$, IoU threshold to $\delta_{\text{IoU}} = 0.3$, modulation weight to $\left[ w_{\text{low}}, w_{\text{high}} \right] = [0.95, 1.05]$. We find that the performance remains fairly robust across datasets when tuning these hyperparameters; thus, we use the same values to report results for all experiments. The detailed findings are provided in the Appendix for reference.

\begin{table*}[]
    \centering 
    \small
    \setlength{\tabcolsep}{5.7mm}{
    \begin{tabular}{l ccc | ccc}
    \toprule
    \multirow{2}{*}{\textbf{Method}} & \multicolumn{3}{c}{\textbf{SA-V val}} & \multicolumn{3}{c}{\textbf{SA-V test}}   \\
     & $\mathcal{J}\&\mathcal{F}$ & $\mathcal{J}$ & $\mathcal{F}$ & $\mathcal{J}\&\mathcal{F}$ & $\mathcal{J}$ & $\mathcal{F}$  \\
    \hline
    SwinB-DeAOT~\citep{yang2022decoupling} & 61.4& 56.6 &66.2 &61.8& 57.2 &66.3\\
    DEVA~\citep{cheng2023tracking} &55.4& 51.5& 59.2& 56.2 &52.4 &60.1 \\
    Cutie-base+~\cite{cheng2024putting} &61.3 &58.3 &64.4& 62.8 &59.8 &65.8\\
    STCN~\citep{cheng2021rethinking} &61.0&  57.4&  64.5&  62.5 & 59.0 & 66.0 \\
    XMem~\citep{cheng2022xmem} & 60.1 &56.3 &63.9& 62.3 &58.9 &65.8 \\\hdashline
    SAM~2~\citep{ravi2024sam} & 76.1 &72.9 &79.2 & 76.0 &72.6 &79.3        \\
    \textbf{SAM2Long (ours)} & 80.8 & 74.7& 84.7 & 80.8 & 76.8 & 84.7\\ 
    SAM 2.1$^\dagger$~\citep{ravi2024sam} & 78.6 &75.1 &82.0 & 79.6 & 76.1 & 83.2        \\
    \textbf{SAM2.1Long (ours)} & 81.1 & 77.5 & 84.7 & 81.2 & 77.6 & 84.9 \\ 
    \bottomrule
    \end{tabular}}
    \caption{Performance comparison with the-state-of-the-arts methods on SA-V dataset.}
    \label{tab:sav_sota}
\end{table*}

\begin{table*}[ht]
    \centering
    \small
    \setlength{\tabcolsep}{3.8mm}{
    \begin{tabular}{l ccc | ccccc}
    \toprule
      \multirow{2}{*}{\textbf{Method}} & \multicolumn{3}{c}{\textbf{LVOS v1}}  & \multicolumn{5}{c}{\textbf{LVOS v2}}   \\
         & $\mathcal{J}\&\mathcal{F}$ & $\mathcal{J}$ & $\mathcal{F}$ & $\mathcal{J}\&\mathcal{F}$ & $\mathcal{J}_s$ & $\mathcal{F}_s$ & $\mathcal{J}_u$  & $\mathcal{F}_u$ \\\hline
         LWL~\citep{bhat2020learning} & 56.4 & 51.8 & 60.9& 60.6 & 58.0& 64.3& 57.2 &62.9\\
         RDE~\citep{li2022recurrent} &53.7& 48.3& 59.2&62.2 &56.7& 64.1 &60.8 &67.2 \\
         STCN~\citep{cheng2021rethinking} & 48.9 &43.9& 54.0 & 60.6 &57.2& 64.0& 57.5 &63.8  \\
         XMem~\citep{cheng2022xmem} & 52.9& 48.1 &57.7& 64.5 & 62.6 & 69.1& 60.6& 65.6 \\
         DDMemory~\citep{hong2023lvos} & 60.7 &55.0 &66.3& - &-&-&-&-\\\hdashline
         SAM~2~\citep{ravi2024sam} & 77.9 &73.1 &82.7& 79.8 & 80.0 & 86.6 & 71.6 & 81.1\\
         \textbf{SAM2Long (ours)} & 81.3 & 76.4 & 86.2 & 84.2 & 82.3  &89.2& 79.1 & 86.2 \\
         SAM 2.1$^\dagger$~\citep{ravi2024sam} & 80.2 & 75.4 & 84.9 & 84.1 & 80.7 & 87.4 & 80.6 & 87.7\\
         \textbf{SAM2.1Long (ours)} & 83.4  & 78.4 & 88.5 & 85.9 & 81.7 & 88.6 & 83.0 & 90.5 \\ 
    \bottomrule
    \end{tabular}}
    \caption{Performance comparison with state-of-the-art methods on validation set of LVOS dataset. Subscript $s$ and $u$ denote scores in seen and unseen categories. Unlike the results presented in Table~\ref{tab:main-results}, we use the evaluation code from the LVOS official repository.}
    \vspace{-3pt}
    \label{tab:lvos}
\end{table*}

\subsection{Main Results}
\textbf{SAM2Long consistently improves SAM~2 over all model sizes and datasets.}
Table~\ref{tab:main-results} presents an overall comparison between SAM~2 and SAM2Long across various model sizes on the SA-V validation and test sets, as well as the LVOS v2 validation set. In total, the table includes 8 model variations, covering SAM 2 and the latest SAM 2.1 across four model sizes.
The average performance across 12 experiments shows an improvement of 3.7 points in the $\mathcal{J} \& \mathcal{F}$ score.
These results confirm that SAM2Long consistently outperforms the SAM~2 baseline by a large margin. For instance, for SAM2Long-Large, achieves an improvement of 4.5 and 5.3 over SAM~2 on SA-V val and test sets. 

\noindent\textbf{SAM2Long outperforms previous methods.} We also compare our proposed method, SAM2Long, with various state-of-the-art VOS methods on both the SA-V~\citep{ravi2024sam} and LVOS~\citep{hong2023lvos, hong2024lvos} datasets, as shown in Table~\ref{tab:sav_sota} and \ref{tab:lvos}. Previous methods such as STCN\cite{cheng2021rethinking} and XMem\cite{cheng2022xmem} adopt memory-based strategies to improve performance and efficiency. STCN models object relations using a single affinity matrix, whereas XMem maintains a three-tier hierarchical memory structure to balance accuracy and GPU usage.  While both methods focus on learning feature-based memory representations, SAM2Long takes a different approach by refining memory at the mask level rather than the feature level. Our method efficiently selects the optimal sequence of memory masks using the proposed tree structure, eliminating the need to retrain the original memory features. As a result, SAM2Long surpasses XMem and STCN by over 20 points on both LVOS v2 and SA-V val, demonstrating its superior ability to handle long-term object segmentation.


\begin{table}[ht]
     \centering
     \small
     \begin{tabular}{lccc}
    \toprule
       \multirow{2}{*}{\textbf{Dataset}}  & \textbf{Duration} & \textbf{SAM~2.1}$^\dagger$ & \textbf{SAM2.1Long} \\
       & seconds & $\mathcal{J}\&\mathcal{F}$ &  $\mathcal{J}\&\mathcal{F}$ \\\hline
       LVOS v1~\cite{hong2023lvos} & 95.4 & 80.2 & 83.4 (\textcolor{red}{3.2$\uparrow$}) \\
       LVOS v2~\cite{hong2024lvos} & 68.4 & 84.1 & 85.9 (\textcolor{red}{1.8$\uparrow$}) \\
       SA-V val~\cite{ravi2024sam} & 13.8 & 78.6 & 81.1 (\textcolor{red}{2.5$\uparrow$}) \\
       SA-V test~\cite{ravi2024sam} & 13.8 & 79.6 & 81.2 (\textcolor{red}{1.6$\uparrow$})\\
       PUMaVOS~\citep{bekuzarov2023xmem++} & 28.7  & 81.1 &  82.4 (\textcolor{red}{1.2$\uparrow$})\\
       VOST~\citep{tokmakov2023breaking} & 21 & 53.1 & 54.0  (\textcolor{red}{0.9$\uparrow$})\\ 
       MOSE~\citep{MOSE} & 12.4 & 74.5  & 75.2 (\textcolor{red}{0.7$\uparrow$}) \\
       YTVOS~\cite{xu2018youtube} & 4.5 & 88.7  & 88.8 (\textcolor{red}{0.1$\uparrow$})\\
       DAVIS-17~\cite{pont20172017} & 1.8 & 90.1  & 90.2 (\textcolor{red}{0.1$\uparrow$})\\
    \bottomrule  
    \end{tabular}
    \caption{The performance comparisons between SAM~2 and SAM2Long on other VOS benchmarks. We list the average duration for each benchmark. All experiments use SAM2.1-L model.}
    \label{tab:other_dataset}
\end{table}
     
\noindent\textbf{Performance gains of SAM2Long positively correlate with video duration.} The results shown in the Table~\ref{tab:other_dataset} reveal a clear trend: SAM2Long tends to exhibit greater performance gains with longer videos.
Although SAM2Long performs comparably to SAM 2 on shorter-duration datasets~\cite{xu2018youtube, pont20172017}, this aligns with the limited long-term tracking opportunities provided by short video sequences. However, SAM2Long can still maintain the strong segmentation performance of SAM 2.


\subsection{Additional Results on VOT Benchmarks}

To further evaluate the effectiveness of SAM2Long, we also apply it to three widely used single-object tracking benchmarks, namely LaSOT~\citep{fan2019lasot}, LaSOText~\citep{fan2021lasot}, and GOT-10k~\citep{huang2019got}. Without whistles and bells, we first generate an initial mask from the bounding box provided for the first frame using the SAM2 Mask decoder. Subsequently, all subsequent masks are transformed into bounding box format by identifying the minimum and maximum coordinates of their non-zero indices.
SAM2Long achieves competitive performance with state-of-the-art VOT methods and SAM 2 baseline, demonstrating its strong generalization ability across various tracking scenarios.

\begin{table}[]
\small
\centering
\begin{tabular}{@{}llll@{}}
\toprule
Dataset & \textbf{LaSoT} & \textbf{LaSoT$_{\text{ext}}$} & \textbf{GoT10k} \\
\midrule
KeepTrack~\citep{mayer2021learning} & 67.1 & 48.2 & - \\
TOMP~\citep{mayer2022transforming} & 68.5 & - & - \\
DropTrack~\citep{wu2023dropmae} & 71.8 & 52.7 & 75.9 \\
SeqTrack~\citep{chen2023seqtrack} & 72.5 & 50.7 & 74.8 \\
MixFormer~\citep{cui2022mixformer} & 70.1 & - & 71.2 \\
GRM-256~\citep{gao2023generalized} & 69.9 & - & 73.4 \\
ROMTrack~\citep{cai2023robust} & 71.4 & 51.3 & 74.2 \\
OSTrack~\citep{ye2022joint} & 71.1 & 50.5 & 73.7 \\
DiffusionTrack~\citep{luo2024diffusiontrack} & 72.3 & - & 74.7 \\
ODTrack~\citep{zheng2024odtrack} & 74.0 & 53.9 & 78.2 \\
LORAT~\citep{lin2025tracking} & 75.1 & 56.6 & 78.2 \\
\midrule
SAM 2~\citep{ravi2024sam} & 70.0 & 56.9 & 80.7 \\
\textbf{SAM2Long (ours)} & 73.9 & 60.9 & 81.1 \\
\bottomrule
\end{tabular}
\caption{Performance comparison with state-of-the-art VOT methods on LaSoT, LaSoT${\text{ext}}$, and GOT-10k benchmarks. The results are reported in terms of AUC (Area Under the success rate Curve) for LaSoT and LaSoT${\text{ext}}$, and AO (Average Overlap) for GOT-10k.}
\end{table}

\begin{figure*}[t!]
    \centering
\includegraphics[width=0.97\linewidth]{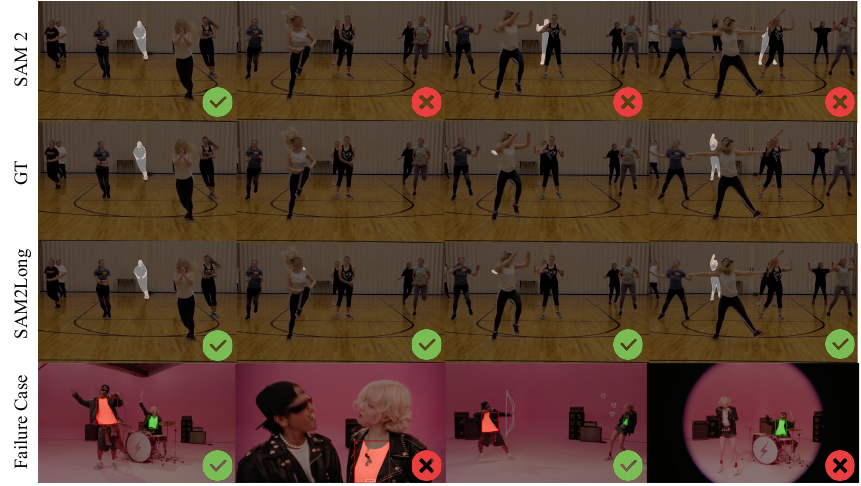}
\vspace{-4pt}
    \caption{Qualitative comparison between SAM~2 and SAM2Long, with GT (Ground Truth) provided for reference. The last row shows a failure case. Best viewed when zoomed in.}
    \label{fig:vis}
    \vspace{-6pt}
\end{figure*}

\subsection{Ablation Study}
We perform a series of ablation studies on the validation subsets of the SA-V and LVOS v2 datasets, utilizing SAM2-Large as the default model size.

\noindent\textbf{Number of Memory Pathways $P$.} We ablate the number of memory pathways to assess their impact on SAM2Long in Table~\ref{tab:memory_number_P}. Note that setting $P=1$ reverts to the SAM~2 baseline. Increasing the number of memory pathways to $P=2$ yields a notable improvement, raising the $\mathcal{J}\&\mathcal{F}$ score to 80.1. This result demonstrates that the proposed memory tree effectively boosts the model's ability to track the correct object while reducing the impact of occlusion. Further increasing the number of memory pathways to $P=3$ achieves the best performance. However, using $P=4$ shows no additional gains, suggesting that three pathways strike the optimal balance between accuracy and computational efficiency for the SAM~2 model.

In terms of speed, maintaining a fixed number of memory pathways at each time step ensures efficiency. Using three pathways incurs only a $14\%$ FPS slowdown, an $8\%$ increase in GFlops, and a $4\%$ rise in memory usage, while boosting performance by 4.5 points on SA-V and 2.4 points on LVOS.

\begin{table}[]
    \centering
    \small
    \begin{tabular}{lccccc}
    \toprule
        $P$  & SA-V & LVOS & FPS & GFlops & Memory \\\hline
        1  & 76.3 & 83.0  & 22 & 844.1 & 5.1GB\\ 
        2   & 80.1 & 85.0 & 21  & 878.2 & 5.2GB \\ 
        \rowcolor{gray!30} 3 & 80.8 & 85.4 & 19 & 912.3 & 5.3GB \\ 
        4   & 80.7 & 85.2 & 17 & 946.4 & 5.4GB\\ 
    \bottomrule
    \end{tabular}
     \caption{Ablation study on number of pathways $P$. We report $\mathcal{J} \& \mathcal{F}$ performance along with FPS, computational cost (GFlops), and memory usage. Throughput and memory consumption are measured on a RTX 3090 GPU with 24 GB of memory.}
    \label{tab:memory_number_P}
\end{table}

\noindent\textbf{Memory Frame Selection.}
Building on IoU-based memory frame selection, we incorporate two additional mechanisms: temporal and spatial selection. The temporal method applies a time decay factor to the IoU score, prioritizing frames closer in time to the current frame. The spatial method integrates frame-wise feature similarity into the IoU score, favoring frames with similar spatial contexts. However, as shown in Table~\ref{tab:memory_frame}, neither approach enhances performance, showing that IoU filtering remains the key factor.

\begin{table}[]
    \centering
    \small
    \begin{tabular}{lcc}
    \toprule
        Memory Selection &  SA-V & LVOS  \\\hline
        \rowcolor{gray!30} Only by IoU &  80.8 & 85.4 \\
        w. temporal & 80.4 & 85.2 \\
        w. spatial & 80.7 & 84.8 \\
    \bottomrule
    \end{tabular}
    \caption{Ablation study on memory frame selection methods. We compare the standard IoU-based selection with additional temporal and spatial selection mechanisms.}
    \label{tab:memory_frame}
\end{table}

\noindent\textbf{Rounding Predicted IoU on Mask Diversity.} 
To evaluate the impact of rounding predicted IoU in selecting diverse masks, we compute the actual IoU between candidate masks chosen based on rounded predicted IoU. As shown in Table~\ref{tab:rounding_places}, rounding to two decimal places significantly reduces the actual IoU, highlighting increased variation among candidates. This enhanced diversity benefits our tree search strategy and improves performance in handling uncertain cases. 

\begin{table}[]
    \centering
    \small
    \begin{tabular}{cccc}
    \toprule
        Rounding &  SA-V & LVOS & actual pairwise IoU \\\hline
        \rowcolor{gray!30} \ding{51} & 80.8  & 85.4 &  51.4 \\
        \ding{53} & 80.4  & 84.4 & 84.5 \\
    \bottomrule
    \end{tabular}
    \caption{Ablation study on rounding predicted IoU.}
    \label{tab:rounding_places}
\end{table}

\subsection{Qualitative Comparison}
We present a qualitative comparison between SAM~2 and SAM2Long in Figure~\ref{fig:vis}. 
As illustrated in Figure~\ref{fig:vis}, SAM 2 successfully tracks the correct person in a group of dancing people. However, when occlusion occurs, SAM~2 mistakenly switches to tracking a different, misleading individual. In contrast, SAM2Long handles this ambiguity effectively. Even during the occlusion, SAM2Long manages to resist the tracking error and correctly resumes tracking the original dancer when she reappears. 
We also present a failure case. When the video features dynamic background changes and distracting elements, SAM2Long struggles to maintain accurate tracking. The model either mistakenly tracks the wrong shirt or completely loses the target. We attribute this to SAM 2's over-reliance on fine-grained visual details and its lack of semantic understanding.

\section{Conclusion}
In this paper, we introduce SAM2Long, a training-free enhancement to SAM~2 that alleviates its limitations in long-term video object segmentation. By employing a constrained tree memory structure with object-aware memory modulation, SAM2Long effectively mitigates error accumulation and improves robustness against occlusions, resulting in a more reliable segmentation process over extended periods. Extensive evaluations on six VOS benchmarks demonstrate that SAM2Long consistently outperforms SAM~2, especially in complex video scenarios. Notably, SAM2Long achieves 
up to a 5-point improvement in $\mathcal{J}\&\mathcal{F}$ scores on challenging long-term video benchmarks such SA-V and LVOS.

\vspace{2pt}
\noindent\textbf{Limitations.} While SAM2Long is effective, it has a few limitations. First, its performance is constrained by the inherent capacity of SAM2, as no learnable parameters are modified, restricting its optimization potential. Second, the method primarily focuses on single-object tracking rather than carefully designed multi-object scenarios. Researching specific multi-object techniques will be a promising future direction.
However, our experiments demonstrate that SAM2Long already performs exceptionally well in both single-object and multi-object cases. We list the comparison in the Appendix.

\clearpage

\section*{Acknowledgemnts}
This work was supported by National Key R\&D Program of China 2022ZD0161600, Shanghai Artificial Intelligence Laboratory, Hong Kong RGC TRS T41-603/20-R, the Centre for Perceptual and Interactive Intelligence (CPII) Ltd under the Innovation and Technology Commission (ITC)’s InnoHK. Dahua Lin is a PI of CPII under the InnoHK.

{
    \small
    \bibliographystyle{ieeenat_fullname}
    \bibliography{main}

\begin{thebibliography}{73}
\providecommand{\natexlab}[1]{#1}
\providecommand{\url}[1]{\texttt{#1}}
\expandafter\ifx\csname urlstyle\endcsname\relax
  \providecommand{\doi}[1]{doi: #1}\else
  \providecommand{\doi}{doi: \begingroup \urlstyle{rm}\Url}\fi

\bibitem[Athar et~al.(2022)Athar, Luiten, Hermans, Ramanan, and Leibe]{athar2022hodor}
Ali Athar, Jonathon Luiten, Alexander Hermans, Deva Ramanan, and Bastian Leibe.
\newblock Hodor: High-level object descriptors for object re-segmentation in video learned from static images.
\newblock In \emph{Proceedings of the IEEE/CVF Conference on Computer Vision and Pattern Recognition}, pages 3022--3031, 2022.

\bibitem[Athar et~al.(2023)Athar, Hermans, Luiten, Ramanan, and Leibe]{athar2023tarvis}
Ali Athar, Alexander Hermans, Jonathon Luiten, Deva Ramanan, and Bastian Leibe.
\newblock Tarvis: A unified approach for target-based video segmentation.
\newblock In \emph{Proceedings of the IEEE/CVF Conference on Computer Vision and Pattern Recognition}, pages 18738--18748, 2023.

\bibitem[Bao et~al.(2018)Bao, Wu, and Liu]{bao2018cnn}
Linchao Bao, Baoyuan Wu, and Wei Liu.
\newblock Cnn in mrf: Video object segmentation via inference in a cnn-based higher-order spatio-temporal mrf.
\newblock In \emph{Proceedings of the IEEE conference on computer vision and pattern recognition}, pages 5977--5986, 2018.

\bibitem[Bekuzarov et~al.(2023)Bekuzarov, Bermudez, Lee, and Li]{bekuzarov2023xmem++}
Maksym Bekuzarov, Ariana Bermudez, Joon-Young Lee, and Hao Li.
\newblock Xmem++: Production-level video segmentation from few annotated frames.
\newblock In \emph{Proceedings of the IEEE/CVF International Conference on Computer Vision}, pages 635--644, 2023.

\bibitem[Bhat et~al.(2020)Bhat, Lawin, Danelljan, Robinson, Felsberg, Van~Gool, and Timofte]{bhat2020learning}
Goutam Bhat, Felix~J{\"a}remo Lawin, Martin Danelljan, Andreas Robinson, Michael Felsberg, Luc Van~Gool, and Radu Timofte.
\newblock Learning what to learn for video object segmentation.
\newblock In \emph{Computer Vision--ECCV 2020: 16th European Conference, Glasgow, UK, August 23--28, 2020, Proceedings, Part II 16}, pages 777--794. Springer, 2020.

\bibitem[Brox and Malik(2010)]{brox2010object}
Thomas Brox and Jitendra Malik.
\newblock Object segmentation by long term analysis of point trajectories.
\newblock In \emph{European conference on computer vision}, pages 282--295. Springer, 2010.

\bibitem[Caelles et~al.(2017)Caelles, Maninis, Pont-Tuset, Leal-Taix{\'e}, Cremers, and Van~Gool]{caelles2017one}
Sergi Caelles, Kevis-Kokitsi Maninis, Jordi Pont-Tuset, Laura Leal-Taix{\'e}, Daniel Cremers, and Luc Van~Gool.
\newblock One-shot video object segmentation.
\newblock In \emph{Proceedings of the IEEE conference on computer vision and pattern recognition}, pages 221--230, 2017.

\bibitem[Cai et~al.(2023)Cai, Liu, Tang, and Wu]{cai2023robust}
Yidong Cai, Jie Liu, Jie Tang, and Gangshan Wu.
\newblock Robust object modeling for visual tracking.
\newblock In \emph{Proceedings of the IEEE/CVF International Conference on Computer Vision}, pages 9589--9600, 2023.

\bibitem[Chen et~al.(2023)Chen, Peng, Wang, Lu, and Hu]{chen2023seqtrack}
Xin Chen, Houwen Peng, Dong Wang, Huchuan Lu, and Han Hu.
\newblock Seqtrack: Sequence to sequence learning for visual object tracking.
\newblock In \emph{Proceedings of the IEEE/CVF conference on computer vision and pattern recognition}, pages 14572--14581, 2023.

\bibitem[Chen et~al.(2018)Chen, Pont-Tuset, Montes, and Van~Gool]{chen2018blazingly}
Yuhua Chen, Jordi Pont-Tuset, Alberto Montes, and Luc Van~Gool.
\newblock Blazingly fast video object segmentation with pixel-wise metric learning.
\newblock In \emph{Proceedings of the IEEE conference on computer vision and pattern recognition}, pages 1189--1198, 2018.

\bibitem[Cheng and Schwing(2022)]{cheng2022xmem}
Ho~Kei Cheng and Alexander~G Schwing.
\newblock Xmem: Long-term video object segmentation with an atkinson-shiffrin memory model.
\newblock In \emph{European Conference on Computer Vision}, pages 640--658. Springer, 2022.

\bibitem[Cheng et~al.(2021)Cheng, Tai, and Tang]{cheng2021rethinking}
Ho~Kei Cheng, Yu-Wing Tai, and Chi-Keung Tang.
\newblock Rethinking space-time networks with improved memory coverage for efficient video object segmentation.
\newblock \emph{Advances in Neural Information Processing Systems}, 34:\penalty0 11781--11794, 2021.

\bibitem[Cheng et~al.(2023)Cheng, Oh, Price, Schwing, and Lee]{cheng2023tracking}
Ho~Kei Cheng, Seoung~Wug Oh, Brian Price, Alexander Schwing, and Joon-Young Lee.
\newblock Tracking anything with decoupled video segmentation.
\newblock In \emph{Proceedings of the IEEE/CVF International Conference on Computer Vision}, pages 1316--1326, 2023.

\bibitem[Cheng et~al.(2024)Cheng, Oh, Price, Lee, and Schwing]{cheng2024putting}
Ho~Kei Cheng, Seoung~Wug Oh, Brian Price, Joon-Young Lee, and Alexander Schwing.
\newblock Putting the object back into video object segmentation.
\newblock In \emph{Proceedings of the IEEE/CVF Conference on Computer Vision and Pattern Recognition}, pages 3151--3161, 2024.

\bibitem[Cox and Hingorani(1996)]{cox1996efficient}
Ingemar~J. Cox and Sunita~L. Hingorani.
\newblock An efficient implementation of reid's multiple hypothesis tracking algorithm and its evaluation for the purpose of visual tracking.
\newblock \emph{IEEE Transactions on pattern analysis and machine intelligence}, 18\penalty0 (2):\penalty0 138--150, 1996.

\bibitem[Cui et~al.(2022)Cui, Jiang, Wang, and Wu]{cui2022mixformer}
Yutao Cui, Cheng Jiang, Limin Wang, and Gangshan Wu.
\newblock Mixformer: End-to-end tracking with iterative mixed attention.
\newblock In \emph{Proceedings of the IEEE/CVF conference on computer vision and pattern recognition}, pages 13608--13618, 2022.

\bibitem[Ding et~al.(2023{\natexlab{a}})Ding, Liu, He, Jiang, Torr, and Bai]{MOSE}
Henghui Ding, Chang Liu, Shuting He, Xudong Jiang, Philip~HS Torr, and Song Bai.
\newblock {MOSE}: A new dataset for video object segmentation in complex scenes.
\newblock In \emph{ICCV}, 2023{\natexlab{a}}.

\bibitem[Ding et~al.(2022)Ding, Xie, Chen, Qian, Zhang, Xiong, and Tian]{ding2022motion}
Shuangrui Ding, Weidi Xie, Yabo Chen, Rui Qian, Xiaopeng Zhang, Hongkai Xiong, and Qi Tian.
\newblock Motion-inductive self-supervised object discovery in videos.
\newblock \emph{arXiv preprint arXiv:2210.00221}, 2022.

\bibitem[Ding et~al.(2023{\natexlab{b}})Ding, Qian, Xu, Lin, and Xiong]{ding2023betrayed}
Shuangrui Ding, Rui Qian, Haohang Xu, Dahua Lin, and Hongkai Xiong.
\newblock Betrayed by attention: A simple yet effective approach for self-supervised video object segmentation.
\newblock \emph{arXiv preprint arXiv:2311.17893}, 2023{\natexlab{b}}.

\bibitem[Duke et~al.(2021)Duke, Ahmed, Wolf, Aarabi, and Taylor]{duke2021sstvos}
Brendan Duke, Abdalla Ahmed, Christian Wolf, Parham Aarabi, and Graham~W Taylor.
\newblock Sstvos: Sparse spatiotemporal transformers for video object segmentation.
\newblock In \emph{Proceedings of the IEEE/CVF conference on computer vision and pattern recognition}, pages 5912--5921, 2021.

\bibitem[Fan et~al.(2019{\natexlab{a}})Fan, Wang, Cheng, and Shen]{fan2019shifting}
Deng-Ping Fan, Wenguan Wang, Ming-Ming Cheng, and Jianbing Shen.
\newblock Shifting more attention to video salient object detection.
\newblock In \emph{Proceedings of the IEEE/CVF conference on computer vision and pattern recognition}, pages 8554--8564, 2019{\natexlab{a}}.

\bibitem[Fan et~al.(2019{\natexlab{b}})Fan, Lin, Yang, Chu, Deng, Yu, Bai, Xu, Liao, and Ling]{fan2019lasot}
Heng Fan, Liting Lin, Fan Yang, Peng Chu, Ge Deng, Sijia Yu, Hexin Bai, Yong Xu, Chunyuan Liao, and Haibin Ling.
\newblock Lasot: A high-quality benchmark for large-scale single object tracking.
\newblock In \emph{Proceedings of the IEEE/CVF conference on computer vision and pattern recognition}, pages 5374--5383, 2019{\natexlab{b}}.

\bibitem[Fan et~al.(2021)Fan, Bai, Lin, Yang, Chu, Deng, Yu, Harshit, Huang, Liu, et~al.]{fan2021lasot}
Heng Fan, Hexin Bai, Liting Lin, Fan Yang, Peng Chu, Ge Deng, Sijia Yu, Harshit, Mingzhen Huang, Juehuan Liu, et~al.
\newblock Lasot: A high-quality large-scale single object tracking benchmark.
\newblock \emph{International Journal of Computer Vision}, 129:\penalty0 439--461, 2021.

\bibitem[Gao et~al.(2023)Gao, Zhou, and Zhang]{gao2023generalized}
Shenyuan Gao, Chunluan Zhou, and Jun Zhang.
\newblock Generalized relation modeling for transformer tracking.
\newblock In \emph{Proceedings of the IEEE/CVF Conference on Computer Vision and Pattern Recognition}, pages 18686--18695, 2023.

\bibitem[Guo et~al.(2025)Guo, Yang, Rao, Meng, Bar-Tal, Ding, Agrawala, Lin, and Dai]{guo2025keyframe}
Yuwei Guo, Ceyuan Yang, Anyi Rao, Chenlin Meng, Omer Bar-Tal, Shuangrui Ding, Maneesh Agrawala, Dahua Lin, and Bo Dai.
\newblock Keyframe-guided creative video inpainting.
\newblock In \emph{Proceedings of the Computer Vision and Pattern Recognition Conference}, pages 13009--13020, 2025.

\bibitem[Hong et~al.(2023)Hong, Chen, Liu, Zhang, Guo, Chen, and Zhang]{hong2023lvos}
Lingyi Hong, Wenchao Chen, Zhongying Liu, Wei Zhang, Pinxue Guo, Zhaoyu Chen, and Wenqiang Zhang.
\newblock Lvos: A benchmark for long-term video object segmentation.
\newblock In \emph{Proceedings of the IEEE/CVF International Conference on Computer Vision}, pages 13480--13492, 2023.

\bibitem[Hong et~al.(2024)Hong, Liu, Chen, Tan, Feng, Zhou, Guo, Li, Chen, Gao, et~al.]{hong2024lvos}
Lingyi Hong, Zhongying Liu, Wenchao Chen, Chenzhi Tan, Yuang Feng, Xinyu Zhou, Pinxue Guo, Jinglun Li, Zhaoyu Chen, Shuyong Gao, et~al.
\newblock Lvos: A benchmark for large-scale long-term video object segmentation.
\newblock \emph{arXiv preprint arXiv:2404.19326}, 2024.

\bibitem[Hu et~al.(2018{\natexlab{a}})Hu, Wang, Kong, Kuen, and Tan]{hu2018motion}
Ping Hu, Gang Wang, Xiangfei Kong, Jason Kuen, and Yap-Peng Tan.
\newblock Motion-guided cascaded refinement network for video object segmentation.
\newblock In \emph{Proceedings of the IEEE conference on computer vision and pattern recognition}, pages 1400--1409, 2018{\natexlab{a}}.

\bibitem[Hu et~al.(2018{\natexlab{b}})Hu, Huang, and Schwing]{hu2018videomatch}
Yuan-Ting Hu, Jia-Bin Huang, and Alexander~G Schwing.
\newblock Videomatch: Matching based video object segmentation.
\newblock In \emph{Proceedings of the European conference on computer vision (ECCV)}, pages 54--70, 2018{\natexlab{b}}.

\bibitem[Huang et~al.(2019)Huang, Zhao, and Huang]{huang2019got}
Lianghua Huang, Xin Zhao, and Kaiqi Huang.
\newblock Got-10k: A large high-diversity benchmark for generic object tracking in the wild.
\newblock \emph{IEEE transactions on pattern analysis and machine intelligence}, 43\penalty0 (5):\penalty0 1562--1577, 2019.

\bibitem[Johnander et~al.(2019)Johnander, Danelljan, Brissman, Khan, and Felsberg]{johnander2019generative}
Joakim Johnander, Martin Danelljan, Emil Brissman, Fahad~Shahbaz Khan, and Michael Felsberg.
\newblock A generative appearance model for end-to-end video object segmentation.
\newblock In \emph{Proceedings of the IEEE/CVF Conference on Computer Vision and Pattern Recognition}, pages 8953--8962, 2019.

\bibitem[Kim et~al.(2015)Kim, Li, Ciptadi, and Rehg]{kim2015multiple}
Chanho Kim, Fuxin Li, Arridhana Ciptadi, and James~M Rehg.
\newblock Multiple hypothesis tracking revisited.
\newblock In \emph{Proceedings of the IEEE international conference on computer vision}, pages 4696--4704, 2015.

\bibitem[Kirillov et~al.(2023)Kirillov, Mintun, Ravi, Mao, Rolland, Gustafson, Xiao, Whitehead, Berg, Lo, et~al.]{kirillov2023segment}
Alexander Kirillov, Eric Mintun, Nikhila Ravi, Hanzi Mao, Chloe Rolland, Laura Gustafson, Tete Xiao, Spencer Whitehead, Alexander~C Berg, Wan-Yen Lo, et~al.
\newblock Segment anything.
\newblock In \emph{Proceedings of the IEEE/CVF International Conference on Computer Vision}, pages 4015--4026, 2023.

\bibitem[Li et~al.(2022)Li, Hu, Xiong, Zhang, Pan, and Liu]{li2022recurrent}
Mingxing Li, Li Hu, Zhiwei Xiong, Bang Zhang, Pan Pan, and Dong Liu.
\newblock Recurrent dynamic embedding for video object segmentation.
\newblock In \emph{Proceedings of the IEEE/CVF Conference on Computer Vision and Pattern Recognition}, pages 1332--1341, 2022.

\bibitem[Li and Loy(2018)]{li2018video}
Xiaoxiao Li and Chen~Change Loy.
\newblock Video object segmentation with joint re-identification and attention-aware mask propagation.
\newblock In \emph{Proceedings of the European conference on computer vision (ECCV)}, pages 90--105, 2018.

\bibitem[Li et~al.(2020)Li, Shen, and Shan]{li2020fast}
Yu Li, Zhuoran Shen, and Ying Shan.
\newblock Fast video object segmentation using the global context module.
\newblock In \emph{Computer Vision--ECCV 2020: 16th European Conference, Glasgow, UK, August 23--28, 2020, Proceedings, Part X 16}, pages 735--750. Springer, 2020.

\bibitem[Liang et~al.(2020)Liang, Li, Jafari, and Chen]{liang2020video}
Yongqing Liang, Xin Li, Navid Jafari, and Jim Chen.
\newblock Video object segmentation with adaptive feature bank and uncertain-region refinement.
\newblock \emph{Advances in Neural Information Processing Systems}, 33:\penalty0 3430--3441, 2020.

\bibitem[Lin et~al.(2025)Lin, Fan, Zhang, Wang, Xu, and Ling]{lin2025tracking}
Liting Lin, Heng Fan, Zhipeng Zhang, Yaowei Wang, Yong Xu, and Haibin Ling.
\newblock Tracking meets lora: Faster training, larger model, stronger performance.
\newblock In \emph{European Conference on Computer Vision}, pages 300--318. Springer, 2025.

\bibitem[Luo et~al.(2024)Luo, Song, Ma, Wei, Yang, and Yang]{luo2024diffusiontrack}
Run Luo, Zikai Song, Lintao Ma, Jinlin Wei, Wei Yang, and Min Yang.
\newblock Diffusiontrack: Diffusion model for multi-object tracking.
\newblock In \emph{Proceedings of the AAAI Conference on Artificial Intelligence}, pages 3991--3999, 2024.

\bibitem[Maninis et~al.(2018)Maninis, Caelles, Chen, Pont-Tuset, Leal-Taix{\'e}, Cremers, and Van~Gool]{maninis2018video}
K-K Maninis, Sergi Caelles, Yuhua Chen, Jordi Pont-Tuset, Laura Leal-Taix{\'e}, Daniel Cremers, and Luc Van~Gool.
\newblock Video object segmentation without temporal information.
\newblock \emph{IEEE transactions on pattern analysis and machine intelligence}, 41\penalty0 (6):\penalty0 1515--1530, 2018.

\bibitem[Mayer et~al.(2021)Mayer, Danelljan, Paudel, and Van~Gool]{mayer2021learning}
Christoph Mayer, Martin Danelljan, Danda~Pani Paudel, and Luc Van~Gool.
\newblock Learning target candidate association to keep track of what not to track.
\newblock In \emph{Proceedings of the IEEE/CVF international conference on computer vision}, pages 13444--13454, 2021.

\bibitem[Mayer et~al.(2022)Mayer, Danelljan, Bhat, Paul, Paudel, Yu, and Van~Gool]{mayer2022transforming}
Christoph Mayer, Martin Danelljan, Goutam Bhat, Matthieu Paul, Danda~Pani Paudel, Fisher Yu, and Luc Van~Gool.
\newblock Transforming model prediction for tracking.
\newblock In \emph{Proceedings of the IEEE/CVF conference on computer vision and pattern recognition}, pages 8731--8740, 2022.

\bibitem[Oh et~al.(2018)Oh, Lee, Sunkavalli, and Kim]{oh2018fast}
Seoung~Wug Oh, Joon-Young Lee, Kalyan Sunkavalli, and Seon~Joo Kim.
\newblock Fast video object segmentation by reference-guided mask propagation.
\newblock In \emph{Proceedings of the IEEE conference on computer vision and pattern recognition}, pages 7376--7385, 2018.

\bibitem[Oh et~al.(2019)Oh, Lee, Xu, and Kim]{oh2019video}
Seoung~Wug Oh, Joon-Young Lee, Ning Xu, and Seon~Joo Kim.
\newblock Video object segmentation using space-time memory networks.
\newblock In \emph{Proceedings of the IEEE/CVF International Conference on Computer Vision}, pages 9226--9235, 2019.

\bibitem[Perazzi et~al.(2016)Perazzi, Pont-Tuset, McWilliams, Van~Gool, Gross, and Sorkine-Hornung]{perazzi2016benchmark}
Federico Perazzi, Jordi Pont-Tuset, Brian McWilliams, Luc Van~Gool, Markus Gross, and Alexander Sorkine-Hornung.
\newblock A benchmark dataset and evaluation methodology for video object segmentation.
\newblock In \emph{Proceedings of the IEEE conference on computer vision and pattern recognition}, pages 724--732, 2016.

\bibitem[Perazzi et~al.(2017)Perazzi, Khoreva, Benenson, Schiele, and Sorkine-Hornung]{perazzi2017learning}
Federico Perazzi, Anna Khoreva, Rodrigo Benenson, Bernt Schiele, and Alexander Sorkine-Hornung.
\newblock Learning video object segmentation from static images.
\newblock In \emph{Proceedings of the IEEE conference on computer vision and pattern recognition}, pages 2663--2672, 2017.

\bibitem[Pont-Tuset et~al.(2017)Pont-Tuset, Perazzi, Caelles, Arbel{\'a}ez, Sorkine-Hornung, and Van~Gool]{pont20172017}
Jordi Pont-Tuset, Federico Perazzi, Sergi Caelles, Pablo Arbel{\'a}ez, Alex Sorkine-Hornung, and Luc Van~Gool.
\newblock The 2017 davis challenge on video object segmentation.
\newblock \emph{arXiv preprint arXiv:1704.00675}, 2017.

\bibitem[Qian et~al.(2023)Qian, Ding, Liu, and Lin]{qian2023semantics}
Rui Qian, Shuangrui Ding, Xian Liu, and Dahua Lin.
\newblock Semantics meets temporal correspondence: Self-supervised object-centric learning in videos.
\newblock In \emph{Proceedings of the IEEE/CVF International Conference on Computer Vision}, pages 16675--16687, 2023.

\bibitem[Qian et~al.(2024)Qian, Ding, and Lin]{qian2024rethinking}
Rui Qian, Shuangrui Ding, and Dahua Lin.
\newblock Rethinking image-to-video adaptation: An object-centric perspective.
\newblock In \emph{European Conference on Computer Vision}, pages 329--348. Springer, 2024.

\bibitem[Ravi et~al.(2024)Ravi, Gabeur, Hu, Hu, Ryali, Ma, Khedr, R{\"a}dle, Rolland, Gustafson, et~al.]{ravi2024sam}
Nikhila Ravi, Valentin Gabeur, Yuan-Ting Hu, Ronghang Hu, Chaitanya Ryali, Tengyu Ma, Haitham Khedr, Roman R{\"a}dle, Chloe Rolland, Laura Gustafson, et~al.
\newblock Sam 2: Segment anything in images and videos.
\newblock \emph{arXiv preprint arXiv:2408.00714}, 2024.

\bibitem[Reid(1979)]{reid1979algorithm}
Donald Reid.
\newblock An algorithm for tracking multiple targets.
\newblock \emph{IEEE transactions on Automatic Control}, 24\penalty0 (6):\penalty0 843--854, 1979.

\bibitem[Robinson et~al.(2020)Robinson, Lawin, Danelljan, Khan, and Felsberg]{robinson2020learning}
Andreas Robinson, Felix~Jaremo Lawin, Martin Danelljan, Fahad~Shahbaz Khan, and Michael Felsberg.
\newblock Learning fast and robust target models for video object segmentation.
\newblock In \emph{Proceedings of the IEEE/CVF conference on computer vision and pattern recognition}, pages 7406--7415, 2020.

\bibitem[Seong et~al.(2020)Seong, Hyun, and Kim]{seong2020kernelized}
Hongje Seong, Junhyuk Hyun, and Euntai Kim.
\newblock Kernelized memory network for video object segmentation.
\newblock In \emph{Computer Vision--ECCV 2020: 16th European Conference, Glasgow, UK, August 23--28, 2020, Proceedings, Part XXII 16}, pages 629--645. Springer, 2020.

\bibitem[Tokmakov et~al.(2023)Tokmakov, Li, and Gaidon]{tokmakov2023breaking}
Pavel Tokmakov, Jie Li, and Adrien Gaidon.
\newblock Breaking the “object” in video object segmentation.
\newblock In \emph{CVPR}, 2023.

\bibitem[Ventura et~al.(2019)Ventura, Bellver, Girbau, Salvador, Marques, and Giro-i Nieto]{ventura2019rvos}
Carles Ventura, Miriam Bellver, Andreu Girbau, Amaia Salvador, Ferran Marques, and Xavier Giro-i Nieto.
\newblock Rvos: End-to-end recurrent network for video object segmentation.
\newblock In \emph{Proceedings of the IEEE/CVF conference on computer vision and pattern recognition}, pages 5277--5286, 2019.

\bibitem[Voigtlaender and Leibe(2017)]{voigtlaender2017online}
Paul Voigtlaender and Bastian Leibe.
\newblock Online adaptation of convolutional neural networks for video object segmentation.
\newblock \emph{arXiv preprint arXiv:1706.09364}, 2017.

\bibitem[Voigtlaender et~al.(2019)Voigtlaender, Chai, Schroff, Adam, Leibe, and Chen]{voigtlaender2019feelvos}
Paul Voigtlaender, Yuning Chai, Florian Schroff, Hartwig Adam, Bastian Leibe, and Liang-Chieh Chen.
\newblock Feelvos: Fast end-to-end embedding learning for video object segmentation.
\newblock In \emph{Proceedings of the IEEE/CVF Conference on Computer Vision and Pattern Recognition}, pages 9481--9490, 2019.

\bibitem[Wang et~al.(2023)Wang, Chen, Wu, Luo, Tang, Dai, Zhao, Xie, Yuan, and Jiang]{wang2023look}
Junke Wang, Dongdong Chen, Zuxuan Wu, Chong Luo, Chuanxin Tang, Xiyang Dai, Yucheng Zhao, Yujia Xie, Lu Yuan, and Yu-Gang Jiang.
\newblock Look before you match: Instance understanding matters in video object segmentation.
\newblock In \emph{Proceedings of the IEEE/CVF conference on computer vision and pattern recognition}, pages 2268--2278, 2023.

\bibitem[Wang et~al.(2019)Wang, Zhang, Bertinetto, Hu, and Torr]{wang2019fast}
Qiang Wang, Li Zhang, Luca Bertinetto, Weiming Hu, and Philip~HS Torr.
\newblock Fast online object tracking and segmentation: A unifying approach.
\newblock In \emph{Proceedings of the IEEE/CVF conference on Computer Vision and Pattern Recognition}, pages 1328--1338, 2019.

\bibitem[Wu et~al.(2023{\natexlab{a}})Wu, Yang, Liu, Wu, Shan, and Chan]{wu2023dropmae}
Qiangqiang Wu, Tianyu Yang, Ziquan Liu, Baoyuan Wu, Ying Shan, and Antoni~B Chan.
\newblock Dropmae: Masked autoencoders with spatial-attention dropout for tracking tasks.
\newblock In \emph{Proceedings of the IEEE/CVF Conference on Computer Vision and Pattern Recognition}, pages 14561--14571, 2023{\natexlab{a}}.

\bibitem[Wu et~al.(2023{\natexlab{b}})Wu, Yang, Wu, and Chan]{wu2023scalable}
Qiangqiang Wu, Tianyu Yang, Wei Wu, and Antoni~B Chan.
\newblock Scalable video object segmentation with simplified framework.
\newblock In \emph{Proceedings of the IEEE/CVF International Conference on Computer Vision}, pages 13879--13889, 2023{\natexlab{b}}.

\bibitem[Xie et~al.(2021)Xie, Yao, Zhou, Zhang, and Sun]{xie2021efficient}
Haozhe Xie, Hongxun Yao, Shangchen Zhou, Shengping Zhang, and Wenxiu Sun.
\newblock Efficient regional memory network for video object segmentation.
\newblock In \emph{Proceedings of the IEEE/CVF conference on computer vision and pattern recognition}, pages 1286--1295, 2021.

\bibitem[Xu et~al.(2018)Xu, Yang, Fan, Yue, Liang, Yang, and Huang]{xu2018youtube}
Ning Xu, Linjie Yang, Yuchen Fan, Dingcheng Yue, Yuchen Liang, Jianchao Yang, and Thomas Huang.
\newblock Youtube-vos: A large-scale video object segmentation benchmark.
\newblock \emph{arXiv preprint arXiv:1809.03327}, 2018.

\bibitem[Yang et~al.(2018)Yang, Wang, Xiong, Yang, and Katsaggelos]{yang2018efficient}
Linjie Yang, Yanran Wang, Xuehan Xiong, Jianchao Yang, and Aggelos~K Katsaggelos.
\newblock Efficient video object segmentation via network modulation.
\newblock In \emph{Proceedings of the IEEE conference on computer vision and pattern recognition}, pages 6499--6507, 2018.

\bibitem[Yang and Yang(2022)]{yang2022decoupling}
Zongxin Yang and Yi Yang.
\newblock Decoupling features in hierarchical propagation for video object segmentation.
\newblock \emph{Advances in Neural Information Processing Systems}, 35:\penalty0 36324--36336, 2022.

\bibitem[Yang et~al.(2020)Yang, Wei, and Yang]{yang2020collaborative}
Zongxin Yang, Yunchao Wei, and Yi Yang.
\newblock Collaborative video object segmentation by foreground-background integration.
\newblock In \emph{European Conference on Computer Vision}, pages 332--348. Springer, 2020.

\bibitem[Yang et~al.(2021{\natexlab{a}})Yang, Wei, and Yang]{yang2021associating}
Zongxin Yang, Yunchao Wei, and Yi Yang.
\newblock Associating objects with transformers for video object segmentation.
\newblock \emph{Advances in Neural Information Processing Systems}, 34:\penalty0 2491--2502, 2021{\natexlab{a}}.

\bibitem[Yang et~al.(2021{\natexlab{b}})Yang, Wei, and Yang]{yang2021collaborative}
Zongxin Yang, Yunchao Wei, and Yi Yang.
\newblock Collaborative video object segmentation by multi-scale foreground-background integration.
\newblock \emph{IEEE Transactions on Pattern Analysis and Machine Intelligence}, 44\penalty0 (9):\penalty0 4701--4712, 2021{\natexlab{b}}.

\bibitem[Ye et~al.(2022)Ye, Chang, Ma, Shan, and Chen]{ye2022joint}
Botao Ye, Hong Chang, Bingpeng Ma, Shiguang Shan, and Xilin Chen.
\newblock Joint feature learning and relation modeling for tracking: A one-stream framework.
\newblock In \emph{European Conference on Computer Vision}, pages 341--357. Springer, 2022.

\bibitem[Zhang et~al.(2023)Zhang, Cui, Wu, and Wang]{zhang2023joint}
Jiaming Zhang, Yutao Cui, Gangshan Wu, and Limin Wang.
\newblock Joint modeling of feature, correspondence, and a compressed memory for video object segmentation.
\newblock \emph{arXiv preprint arXiv:2308.13505}, 2023.

\bibitem[Zhang et~al.(2019)Zhang, Lin, Zhang, Lu, and He]{zhang2019fast}
Lu Zhang, Zhe Lin, Jianming Zhang, Huchuan Lu, and You He.
\newblock Fast video object segmentation via dynamic targeting network.
\newblock In \emph{Proceedings of the IEEE/CVF International Conference on Computer Vision}, pages 5582--5591, 2019.

\bibitem[Zheng et~al.(2024)Zheng, Zhong, Liang, Mo, Zhang, and Li]{zheng2024odtrack}
Yaozong Zheng, Bineng Zhong, Qihua Liang, Zhiyi Mo, Shengping Zhang, and Xianxian Li.
\newblock Odtrack: Online dense temporal token learning for visual tracking.
\newblock In \emph{Proceedings of the AAAI Conference on Artificial Intelligence}, pages 7588--7596, 2024.

\bibitem[Zhou et~al.(2024)Zhou, Pang, and Wang]{zhou2024rmem}
Junbao Zhou, Ziqi Pang, and Yu-Xiong Wang.
\newblock Rmem: Restricted memory banks improve video object segmentation.
\newblock In \emph{Proceedings of the IEEE/CVF Conference on Computer Vision and Pattern Recognition}, pages 18602--18611, 2024.

\end{thebibliography}
}

\clearpage

\appendix

\section{Dataset Information}
\noindent\textbf{SA-V}~\citep{ravi2024sam} is a large-scale video segmentation dataset designed for promptable visual segmentation across diverse scenarios. It encompasses 50.9K video clips, aggregating to 642.6K masklets with 35.5M meticulously annotated masks. The dataset presents a challenge with its inclusion of small, occluded, and reappearing objects throughout the videos. The dataset is divided into training, validation, and testing sets, with most videos allocated to the training set for robust model training. The validation set has 293 masklets across 155 videos for model tuning, while the testing set includes 278 masklets across 150 videos for comprehensive evaluation.

\vspace{2pt}
\noindent\textbf{LVOS v1}~\citep{hong2023lvos} is a VOS benchmark for long-term video object segmentation in realistic scenarios. It comprises 720 video clips with 296,401 frames and 407,945 annotations, with an average video duration of over 60 seconds. LVOS introduces challenging elements such as long-term object reappearance and cross-temporal similar objects.  In LVOS v1, the dataset includes 120 videos for training, 50 for validation, and 50 for testing. 

\vspace{2pt}
\noindent\textbf{LVOS v2}~\citep{hong2024lvos} expends LVOS v1 and provides 420 videos for training, 140 for validation, and 160 for testing. This paper primarily utilizes v2, as it already includes the sequences present in v1. The dataset spans 44 categories, capturing typical everyday scenarios, with 12 of these categories deliberately left unseen to evaluate and better assess the generalization capabilities of VOS models.

\vspace{2pt}
\noindent\textbf{MOSE}~\citep{MOSE} is a challenging VOS dataset targeted on complex, real-world scenarios, featuring 2,149 video clips with 431,725 high-quality segmentation masks. These videos are split into 1,507 training videos, 311 validation videos, and 331 testing videos. 

\vspace{2pt}
\noindent\textbf{VOST}~\citep{tokmakov2023breaking} is a semi-supervised video object segmentation benchmark that emphasizes complex object transformations. Unlike other datasets, VOST includes objects that are broken, torn, or reshaped, significantly altering their appearance. It comprises more than 700 high-resolution videos, captured in diverse settings, with an average duration of 21 seconds, all densely labeled with instance masks.

\vspace{2pt}
\noindent\textbf{PUMaVOS}~\citep{bekuzarov2023xmem++} is a novel video dataset designed for benchmarking challenging segmentation tasks. It includes 24 video clips, each ranging from 13.5 to 60 seconds (28.7 seconds on average) at 480p resolution with varying aspect ratios. PUMaVOS focuses on difficult scenarios where annotation boundaries do not align with clear visual cues, such as half faces, necks, tattoos, and pimples, commonly encountered in video production.

\vspace{2pt}
\noindent\textbf{YouTubeVOS-2019}~\citep{xu2018youtube} is a large-scale video object segmentation dataset featuring 3,252 sequences with detailed annotations at 6 FPS across 78 diverse categories, including humans, animals, vehicles, and accessories. Each video clip is between 3 to 6 seconds long and frequently contains multiple objects, which have been manually segmented by professional annotators.

\vspace{2pt}
\noindent\textbf{DAVIS2017}~\citep{pont20172017} is a well-known benchmark dataset comprising 60 training videos and 30 validation videos, with a total of 6,298 frames. It offers high-quality, pixel-level annotations for every frame, making it a standard resource for evaluating different VOS methods.

\vspace{2pt}
\noindent\textbf{LaSOT}~\citep{fan2019lasot} is a large-scale tracking dataset designed for long-term visual object tracking. It contains 1,400 videos spanning 70 object categories, with an average sequence length exceeding 2,500 frames, making it a challenging benchmark for evaluating tracking algorithms. \textbf{LaSOText}\citep{fan2021lasot} is an extension of LaSOT, featuring a subset of 15 categories with 150 videos.

\vspace{2pt}
\noindent\textbf{GOT-10k}~\citep{huang2019got} is a large-scale generic object tracking dataset that covers over 10,000 video sequences, spanning more than 560 object classes. It provides strict one-shot evaluation settings and diverse real-world tracking scenarios, making it a widely used benchmark for developing and assessing tracking models.

\begin{table*}[htbp]
    \centering
    \begin{minipage}{\textwidth}
        \centering
        \small
        \begin{tabular}{ccccc>{\columncolor{gray!30}}ccccccc}
        \toprule
            $\delta_{\text{iou}}$ & SAM 2 & 0 & 0.1& 0.2& 0.3& 0.4& 0.5& 0.6& 0.7& 0.8 & 0.9\\\hline
            SA-V  & 76.3 & 80.0& 80.7& 80.6 &80.8& 80.7& 81.0& 80.6& 80.8& 80.0& 77.8 \\
            LVOS  & 83.0 & 84.0& 85.1& 85.6& 85.4& 85.4&85.1& 85.6& 85.2& 84.1 & 83.5 \\
        \bottomrule
        \end{tabular}
        \caption{Ablation study on IoU threshold $\delta_{\text{iou}}$.}
        \label{tab:iou}
    \end{minipage}

    \vspace{0.3cm}

    \begin{minipage}{\textwidth}
        \centering
        \small
        \begin{tabular}{cccc>{\columncolor{gray!30}}cccc}
        \toprule
            $\delta_{\text{conf}}$ & SAM 2 & 0 & 1 & 2& 3& 4& 5\\\hline
            SA-V  & 76.3  & 80.4 & 80.4 & 80.8 &  80.8 & 79.7 & 78.8  \\
            LVOS  & 83.0 &  84.4 & 84.6  & 85.4 & 85.2 & 84.1 & 83.9 \\
        \bottomrule
        \end{tabular}
        \caption{Ablation study on uncertainty threshold $\delta_{\text{conf}}$.}
        \label{tab:Uncertainty}
    \end{minipage}

    \vspace{0.3cm}

    \begin{minipage}{\textwidth}
        \centering
        \small
        \begin{tabular}{cccccc>{\columncolor{gray!30}}cc}
        \toprule
            $[ w_{\text{low}}, w_{\text{high}}]$ & SAM 2 &  $[0.6,1.4]$   &$[0.7,1.3]$   & $[0.8,1.2]$   & $[0.9,1.1]$  & $[0.95,1.05]$  & $[1,1]$ \\\hline
            SA-V  & 76.3 &  77.1&  79.3&  79.8&  80.4&  80.8&  80.5 \\
            LVOS  & 83.0 & 77.0& 78.8& 82.9& 85.0& 85.4& 84.9\\
        \bottomrule
        \end{tabular}
        \caption{Ablation study on modulation weight $[ w_{\text{low}}, w_{\text{high}}]$.}
        \label{tab:modulation_weights}
    \end{minipage}

    \vspace{0.3cm}
     \begin{minipage}{\textwidth}
     \centering
     \small
     \begin{tabular}{llll}
    \toprule
       \textbf{SA-V val}  &  Single-object & Multi-object & Overall \\
       \# of seq & 73 & 82 & 155 \\\hline
       SAM 2.1 & 79.0 & 78.4 &  78.6 \\
       SAM2.1Long  & 80.6 (\textcolor{red}{1.6\% $\uparrow$}) & 81.3 (\textcolor{red}{2.9\% $\uparrow$}) & 81.1 (\textcolor{red}{2.5\% $\uparrow$}) \\
       \bottomrule
    \end{tabular}
        \caption{Performance comparison on single-object sequence and multi-object sequence. SAM2Long performs exceptionally well in both single-object and multi-object cases.}
    \label{tab:multi-object}
     \end{minipage}
     \vspace{-5pt}
\end{table*}

\section{More Ablation Study}
\noindent\textbf{Iou Threshold $\delta_{\text{iou}}$.} The choice of the IoU threshold $\delta_{\text{iou}}$ is crucial for selecting frames with reliable object cues. As shown in Table~\ref{tab:iou}, setting $0.1 \geq \delta_{\text{iou}}  \geq 0.7$ yields the competitive $\mathcal{J}\&\mathcal{F}$, indicating an effective trade-off between filtering out poor-quality frames and retaining valuable segmentation information. In contrast, setting no quality requirement for masks ($\delta_{\text{iou}} = 0$) lowers the score to 80.0, as unreliable frames with poor segmentation harm SAM 2. Conversely, an overly strict selection ($\delta_{\text{iou}} \geq 0.8$) further degrades performance by excluding important neighboring frames, forcing the model to rely on distant frames as memory.

\noindent\textbf{Uncertainty Threshold $\delta_{\text{conf}}$.} The uncertainty threshold $\delta_{\text{conf}}$ controls the selection of hypotheses under uncertain conditions. Our results in Table~\ref{tab:Uncertainty} indicate that setting $\delta_{\text{conf}}$ to 2 provides the highest $\mathcal{J}\&\mathcal{F}$ score, indicating an optimal level for uncertainty handling. Lower values (e.g., 0) lead to suboptimal performance by committing to incorrect segmentations, causing error propagation. Higher values (e.g., 4) do not improve performance, indicating that beyond a certain threshold, the model efficiently relies on top-scoring masks without needing additional diversity.

\noindent\textbf{Memory Attention Modulation $[ w_{\text{low}}, w_{\text{high}}]$.} We explore the effect of modulating the attention weights for memory entries using different ranges in Table~\ref{tab:modulation_weights}. The configuration $\left[1, 1\right]$ means no modulation is applied. We find that the configuration of $\left[0.95, 1.05\right]$ achieves the best performance while increasing the modulation range decreases performance. This result indicates that slight modulation sufficiently emphasizes reliable memory entries.

\section{More Quantitive Comparion}
In this section, we present a detailed comparison of SAM2.1 and SAM2Long across single-object and multi-object sequences, as shown in Table~\ref{tab:multi-object}. Our experiments demonstrate that SAM2Long outperforms SAM2.1 in both single-object and multi-object scenarios. Specifically, SAM2Long achieves a 1.6\% improvement in single-object sequences, a 2.9\% improvement in multi-object sequences, and an overall 2.5\% enhancement in performance. These results highlight SAM2Long's robustness and effectiveness in various video segmentation tasks.

\section{More Visualization}
\begin{figure*}[t!]
    \centering
    \includegraphics[width=0.85\linewidth]{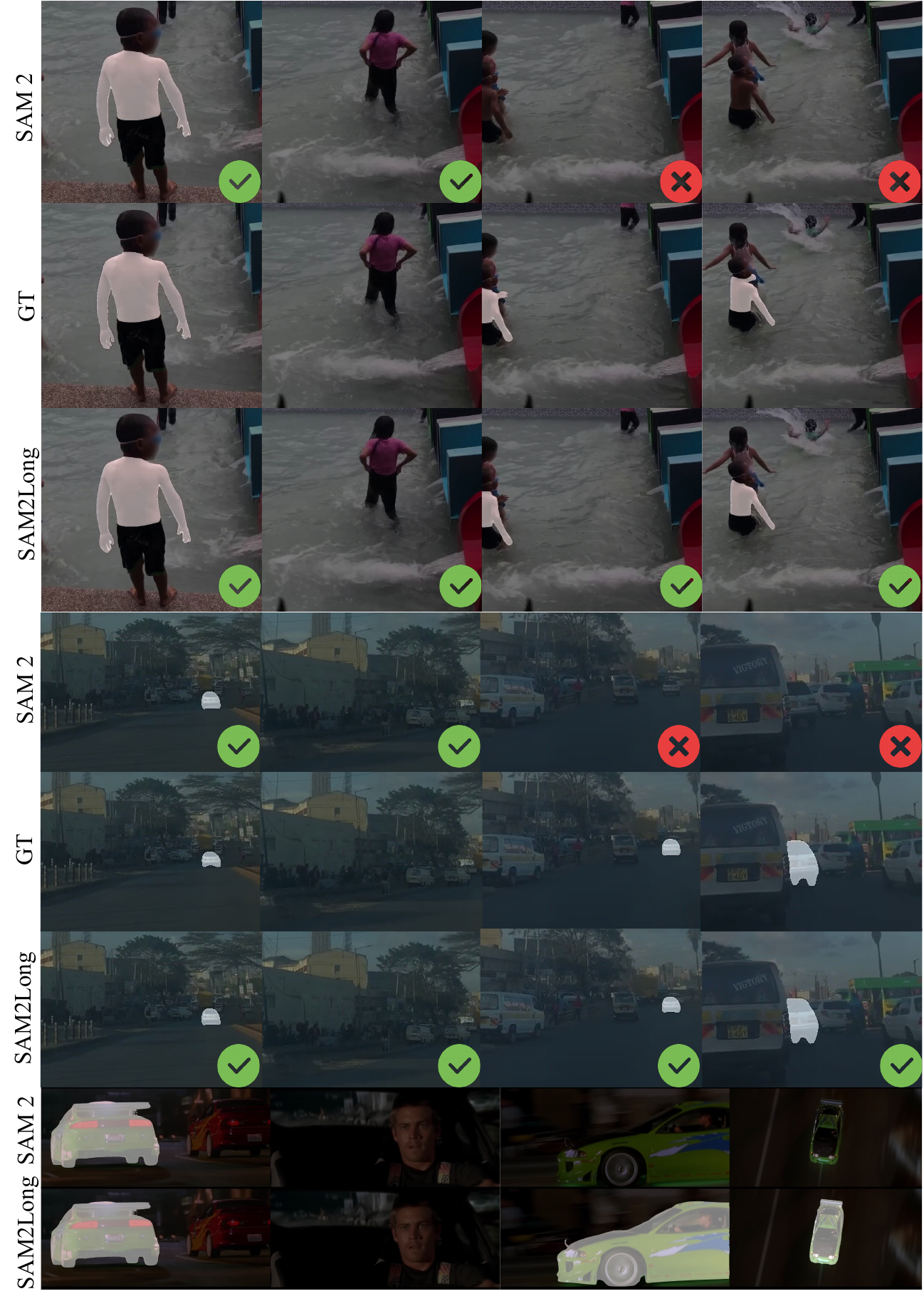}
    \caption{Qualitative comparison between SAM~2 and SAM2Long, with GT (Ground Truth) provided for reference. The last row shows an in-wild case. Best viewed when zoomed in.}
    \label{fig:vis_app}
\end{figure*}

We present additional comparisons between SAM2 and SAM2Long in Figure~\ref{fig:vis_app}. SAM2Long significantly reduces segmentation errors, showing improved accuracy and consistency in object tracking across frames. Notably, in the Fast \& Furious movie scene, SAM2Long successfully tracks the green car, even under challenging dynamic camera movements. Overall, SAM2Long offers substantial improvements over SAM2, especially in handling object occlusion and reappearance, leading to better performance in long-term video segmentation tasks.



\end{document}